\documentclass[runningheads]{llncs}
\usepackage[T1]{fontenc}
\usepackage{graphicx,verbatim}
\usepackage{subcaption}
\usepackage{hyperref}
\usepackage{booktabs}
\usepackage[margin=1in]{geometry}
\usepackage{array}
\usepackage{float}
\usepackage{subcaption}
\begin{document}
\title{MedPAO: A Protocol-Driven Agent for Structuring Medical Reports}
%
%
\author{Shrish Shrinath Vaidya\inst{1}\orcidID{0009-0006-5438-9500} \and
Gowthamaan Palani\inst{2}\orcidID{0009-0009-4095-2413} \and
Sidharth Ramesh\inst{1}\orcidID{0000-0001-5172-5739} \and
Velmurugan Balasubramanian\inst{3} \and
Minmini Selvam\inst{4} \and
Gokulraja Srinivasaraja\inst{5} \and
Ganapathy Krishnamurthi\inst{1}
}
%

\authorrunning{Vaidya et al.}
%
\institute{Department of Data Science and AI, IIT Madras, India \and
Department of Engineering Design, IIT Madras, India \and
LoveForm Health Technologies, India \and
Department of Radiology and Imaging Sciences, Sri Ramachandra Institute of Higher Education and Research, India \and
Department of Neuro and Interventional Radiology, Sri Ramachandra Institute of Higher Education and Research, India\\
\email{shrish.s.vaidya@dsai.iitm.ac.in}\\}

%
\maketitle              
\begin{abstract}
The deployment of Large Language Models (LLMs) for structuring clinical data is critically hindered by their tendency to hallucinate facts and their inability to follow domain-specific rules. To address this, we introduce MedPAO, a novel agentic framework that ensures accuracy and verifiable reasoning by grounding its operation in established clinical protocols such as the ABCDEF protocol for CXR analysis. MedPAO decomposes the report structuring task into a transparent process managed by a Plan-Act-Observe (PAO) loop and specialized tools. This protocol-driven method provides a verifiable alternative to opaque, monolithic models. The efficacy of our approach is demonstrated through rigorous evaluation: MedPAO achieves an F1-score of 0.96 on the critical sub-task of concept categorization. Notably, expert radiologists and clinicians rated the final structured outputs with an average score of 4.52 out of 5, indicating a level of reliability that surpasses baseline approaches relying solely on LLM-based foundation models. The code is available at: \url{https://github.com/MiRL-IITM/medpao-agent}

\end{abstract}
\section{Introduction}
Clinical narratives like radiology reports contain a wealth of observational data, but their unstructured, free-text format hinders large-scale computational analysis \cite{kreimeyer2017natural,bose2021survey}. Radiologists often use systematic checklists, such as the ABCDEF mnemonic for chest X-rays \cite{Jones_etal_CXR_2025}, to ensure diagnostic completeness. The \textbf{ABCDEF} protocol is a mnemonic-driven, structured framework for systematic chest X-ray interpretation. It organizes key anatomical regions and potential abnormalities into six distinct categories to promote thorough and consistent reporting. \textbf{A} evaluates airways, \textbf{B} assesses lung fields and pleura, \textbf{C} examines cardiomediastinal contours and great vessels, and \textbf{D} focuses on the diaphragm and subdiaphragmatic structures. \textbf{E} reviews external anatomy including bones and soft tissues, while \textbf{F} identifies foreign bodies and medical devices. However, this structure is typically lost in the final dictated report, leading to significant inter-clinician variability that complicates data aggregation and the development of predictive models \cite{donnelly2019using,nobel2022structured}.

\begin{figure}[H]
    \centering
    \includegraphics[width=0.90\linewidth]{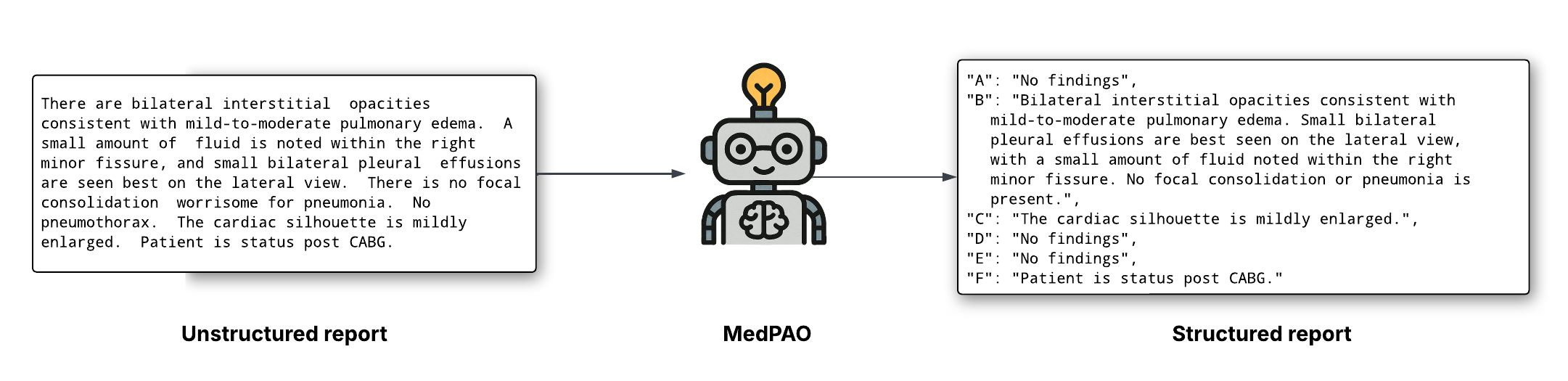}
    \caption{Illustrative example of MedPAO}
    \label{fig:intro}
\end{figure}
\begin{figure}
    \centering
    \includegraphics[width=1\linewidth]{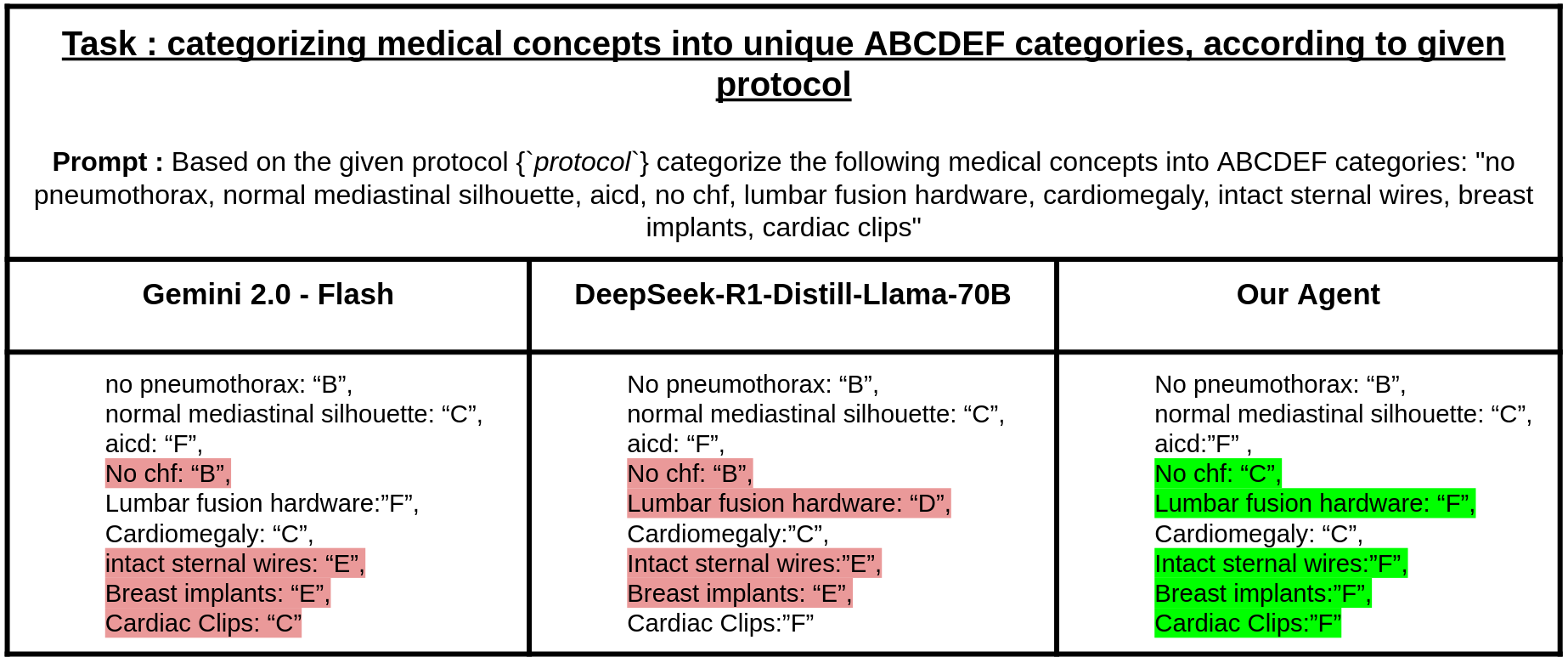}
    \caption{Proposed agent vs SOTA LLMs on medical concept categorization task according to ABCDEF protocol}
    \label{fig1}
\end{figure}

Existing approaches to automate data structuring have significant drawbacks. Traditional rule-based and supervised machine learning methods are either too brittle or require prohibitively expensive manual annotation \cite{aggarwal2018hedea,bose2021survey}. While modern Large Language Models (LLMs) show impressive zero-shot capability, they are prone to factual "hallucination" and lack verifiable reasoning essential for high-stakes medical applications \cite{huang2024critical,alzaid2024large,ntinopoulos2025large}. This creates a critical need for a framework that is both flexible and trustworthy. To address this challenge, we introduce MedPAO, a Protocol-Driven Agent for Structuring Medical Reports. Figure \ref{fig:intro} illustrates MedPAO's structured output on given report findings. MedPAO is a novel agentic framework that operationalizes a clinical protocol as its core reasoning structure.  Our agent employs a Plan-Act-Observe (PAO) loop \cite{yao2023react} to orchestrate specialized, verifiable tools, systematically mapping narrative findings back to a clinical protocol like the ABCDEF system. This approach creates a consistent, structured representation of the report's content. As shown in Figure \ref{fig1}, our agent correctly categorizes medical concepts where other LLMs fail, demonstrating a clear path towards reliable and transparent report structuring.

\noindent The primary contributions of this work are:
\begin{itemize}
\item We propose MedPAO, a novel, zero-shot agentic framework that transforms unstructured medical reports into protocol-compliant structured data, standardizing information across different clinician writing styles.
\item We demonstrate the efficacy of our protocol-driven approach by implementing the ABCDEF protocol, showing how our agent maps narrative findings to a clinically established standard.
\item We introduce a modular and verifiable toolset, including a fine-tuned medical concept extractor and reasoning-based modules for ontology filtering and protocol categorization.
\item Through rigorous evaluation, we show that MedPAO significantly outperforms baseline methods and provides a scalable, transparent, and accurate solution for structuring medical reports.
\end{itemize}

\section{Related Works}
\label{sec:related_works}
The effort to structure clinical text has a long history, beginning with brittle rule-based systems that failed to generalize across linguistic variations \cite{aggarwal2018hedea,kreimeyer2017natural,zeng2019rich}. This led to a shift towards supervised machine learning, which offered more flexibility but introduced a significant "annotation bottleneck" due to the high cost of creating expert-labeled datasets \cite{spasic2020clinical,srivastava2025medpromptextract}.

The emergence of Large Language Models (LLMs) marked a new paradigm, promising zero-shot information extraction in varied domains from medicine to historical OCR-processed texts \cite{ntinopoulos2025large,grothey2025comprehensive,schwitter2025using}. However, monolithic LLMs act as opaque "black boxes" \cite{de2023structuring}, are prone to factual "hallucinations" and often lack the deep domain knowledge required to correctly interpret complex clinical rules like TNM staging \cite{huang2024critical}. This core structuring task is distinct from parallel work in automated report generation, which produces more unstructured text \cite{WANG2023100033,hyland2024}, or specialized evaluation frameworks whose schemas are not designed for general-purpose data standardization \cite{jiang2025}.

To enhance LLM reliability, a new generation of sophisticated frameworks has been developed. One class of solutions focuses on post-hoc validation, using two-stage pipelines for confidence scoring \cite{alzaid2024large} or self-correcting feedback loops for error checking \cite{bisercic2023interpretable}. Another approach involves multi-agent architectures, which follow distinct strategies. Some employ a consensus mechanism, where one LLM adjudicates the outputs of several others to improve robustness \cite{tripathi2025employing}. In contrast, others use a delegation model, as seen in the financial domain, where specialized agents handle discrete sub-tasks like extraction and querying \cite{choi2025structuring}. A third strategy involves fine-tuning LLMs on annotated data to directly output structured formats like JSON, though this re-introduces the annotation bottleneck \cite{dagdelen2024structured}.
While these state-of-the-art methods improve reliability through correction, consensus, or delegation, they do not ground the reasoning process within an explicit, verifiable clinical workflow. This leaves a critical gap for a system whose internal logic is both transparent and clinically valid. MedPAO addresses this gap by uniquely operationalizing an established medical protocol as the core, step-by-step reasoning structure of an agentic framework, ensuring that the final structured output is accurate and demonstrably aligned with standard clinical practice.

\section{Methodology}
\subsection{Overview of Agentic Framework}    

\begin{figure}[!htbp]
    \centering
    \includegraphics[width=1\linewidth]{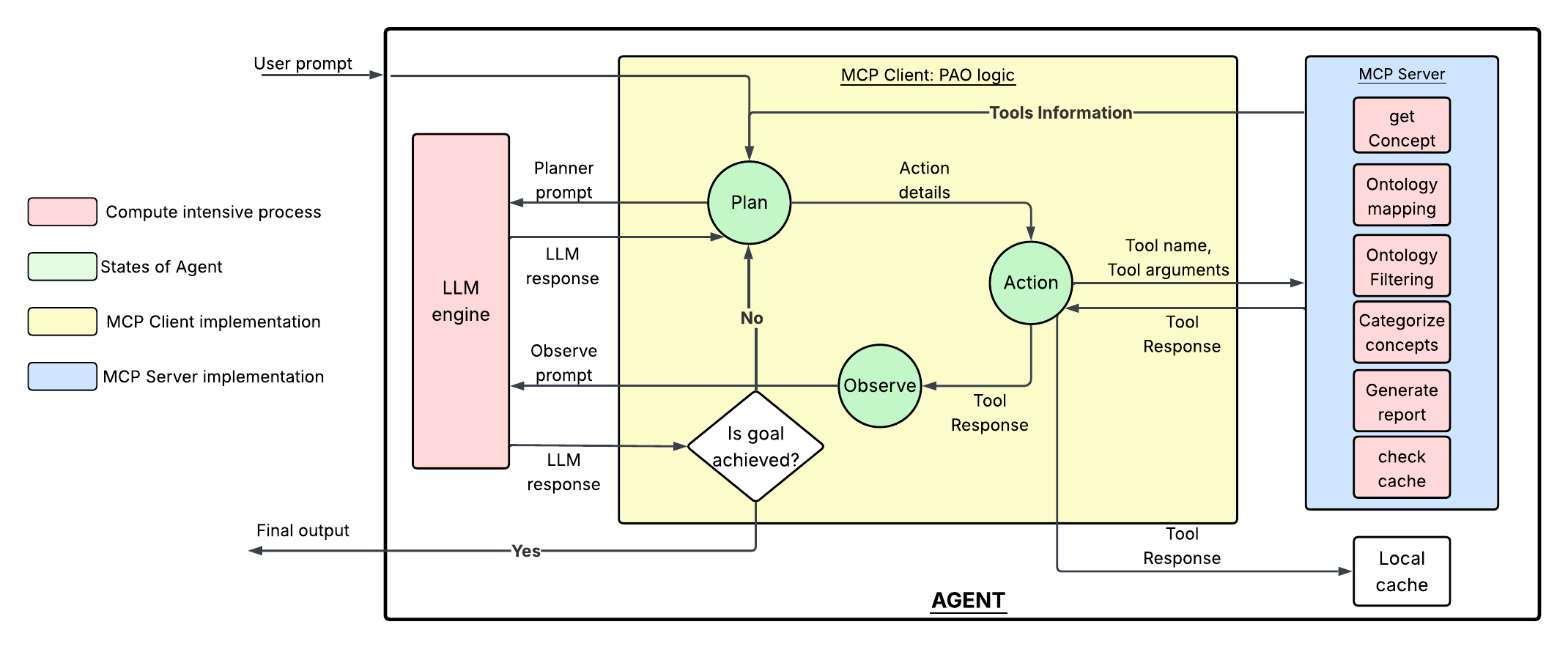}
    \caption{Architecture of proposed MedPAO agent for medical report processing tasks.}
    \label{fig2}
\end{figure}

To facilitate structured report generation, our framework integrates a Model Context Protocol (MCP) with a Plan-Act-Observe (PAO) loop, enabling dynamic orchestration of auxiliary tools in conjunction with a central large language model (LLM) engine, as depicted in Fig. \ref{fig2}. The PAO loop not only determines which tools to invoke based on user input, but also provides the full contextual scope to the LLM, ensuring that the LLM maintains overarching control over the decision-making process.

\subsection{Agent Architecture and Implementation}

\subsubsection{LLM engine}
Our agent employs the foundation model, Deepseek-R1-70B \cite{deepseekr1}, as its core LLM engine. This selection was driven by two key factors: (1) the model's extended context window capacity, and (2) its state-of-the-art reasoning performance. The architectural features of the model serves a dual purpose: it maintains output discipline while simultaneously providing model explainability.

\subsubsection{The PAO logic and MCP}

Our framework integrates the complete Plan-Act-Observe (PAO) framework with the Model-Context Protocol (MCP), a standardized approach for developing efficient AI agents. The complementary MCP client implements our PAO logic to orchestrate the agentic workflow. Our client strategically employs the LLM engine during both the Plan and Observe states of the agent's operational cycle. In the Plan state, the MCP client prompts the LLM engine by providing contextual information including available tools and any previously executed steps to determine the next appropriate tool along with its specific parameters.  The server then executes the designated tools with the provided parameters and returns structured outputs. During the Observe state, the system presents the LLM with both the tool-generated output and the original user query, requesting an evaluation on whether the response adequately addresses the initial request. Based on this assessment, the LLM decides whether to continue iterating through the cycle or terminate the process and return the current output as final. This modular client-server design provides benefits in terms of both efficiency and scalability.

\subsubsection{The Agent's Toolset}

 \paragraph{Tool 1: get Concept:}
 The Concept Extraction Module serves as the foundational component of our toolset, specifically designed to identify and extract clinically relevant medical terms and diagnostic findings from unstructured medical reports. In our framework, "concepts" encompass both pathological observations and their associated anatomical locations when specified. The module utilizes a fine-tuned MedLlama-8B\cite{johnsnowlabs_medllama3_2024} model for medical concept recognition. Due to the absence of existing ground truth report-concept pairs, we engaged a certified radiologist and clinician to annotate a preliminary set of 40 Chest X-ray reports from the Eye Gaze MIMIC-CXR dataset \cite{karargyris2021creation}. Leveraging these expert-validated samples, we generated additional synthetic training data using foundation models like GPT-4, Gemini 2.5-Flash, Claude 3.5 and Deepseek-R1 through few-shot prompting techniques and restricting them to a predefined vocabulary from SNOMEDCT and RADLEX ontologies. From each of the foundation models we extracted 40 synthetic concept-report pairs, resulting in a final dataset comprising of 200 unique samples with 311 unique medical concepts. The MedLlama model was fine-tuned using LoRA (Low-Rank Adaptation) with full precision (non-quantized) parameters on the mixture of synthetic and original data. During inference, we implemented a few-shot prompting approach to extract both medical concepts and their corresponding source sentences with the fine-tuned model, ensuring traceability and reducing extraction errors.

\paragraph{Tool 2: Ontology Mapping: }
To enhance the clinical context of extracted medical concepts,we incorporated well-established medical ontologies—specifically SNOMED-CT and RADLEX—using the BioPortal Annotator API\cite{martinez2017ncbo}, as they offer comprehensive coverage and promote interoperability for a wide range of radiological findings. This module operates downstream of the concept extraction tool, automatically mapping each identified medical concept to its corresponding ontological representation. For every extracted medical concept, the API retrieves the precise ontological mapping as well as its complete ancestral hierarchy, thereby providing rich, structured clinical context. Lemmatization was done on the raw concepts as the ontology annotator tool does only rule based matching. The retrieved ontological mappings were cached locally for efficiency and subsequently fed into the LLM pipeline to support downstream reasoning tasks. 

\paragraph{Tool 3: Ontology filtering: }
In our framework, medical concepts are typically expressed as multi-word terms, which often results in multiple potential ontology mappings for each concept. To address this complexity, we developed a systematic approach to classify these mappings into \textit{Primary} and \textit{Secondary} ontologies. The Primary Ontology represents the principal pathological finding or clinical abnormality, while Secondary Ontologies encompass supplementary details such as severity, anatomical location, or associated observations. 

The need for such segregation arises from a key challenge that, a single concept may map to multiple ontologies with distinct semantic meanings, leading to potential ambiguity.  To automate this classification, we leverage the LLM engine, which uses a few-shot prompting strategy.

\paragraph{Tool 4: Categorize concepts: }
Our framework incorporates a dedicated tool for protocol-based categorization of medical concepts, essential for transforming unstructured clinical findings into standardized reports. For this study, we implemented the  "ABCDEF" protocol \cite{Jones_etal_CXR_2025} for chest X-ray diagnosis, which systematically classifies findings based on anatomical regions and abnormal findings into six categories (A-F). Each extracted medical concept is precisely mapped to its corresponding anatomical category, enabling the conversion of free-text observations into a structured format while maintaining clinical accuracy. This approach not only addresses the variability inherent in traditional radiology reporting but also establishes a foundation for consistent, protocol-compliant documentation.

For this task, we leverage the LLM engine to process multiple inputs: (1) the complete protocol description \cite{Jones_etal_CXR_2025}, (2) the extracted medical concepts, and (3) their corresponding filtered ontologies. The ontological information serves as critical contextual data, enabling more accurate and clinically relevant categorization of each concept according to the specified protocol.

\paragraph{Tool 5: generate Report: }
Our framework generates protocol-compliant structured findings by processing categorized medical concepts with the LLM engine. The model receives an input, comprising the complete protocol specifications alongside the classified concepts and their original source sentences. This approach ensures the generated findings are clinically accurate, grounded in the source report's terminology, and adhere to the required format. The selection of a foundational model was necessitated by the absence of structured ground truth data for supervised training. We leveraged its inherent generalization capabilities to avoid the performance degradation in text generation, on unseen data distributions commonly observed with fine-tuning\cite{wang2024twostagellmfinetuningspecialization,luo2025empiricalstudycatastrophicforgetting}.

\paragraph{Tool 6: check cache: }
To optimize computational efficiency, our framework caches recurring medical concepts and their protocol-specific categorizations. Reusing these pre-processed mappings from local storage accelerates report generation and reduces resource utilization without compromising accuracy. This caching strategy is vital for meeting the performance demands for clinical deployment.

\begin{figure}[!htbp]
    \centering
    \includegraphics[width=1\linewidth]{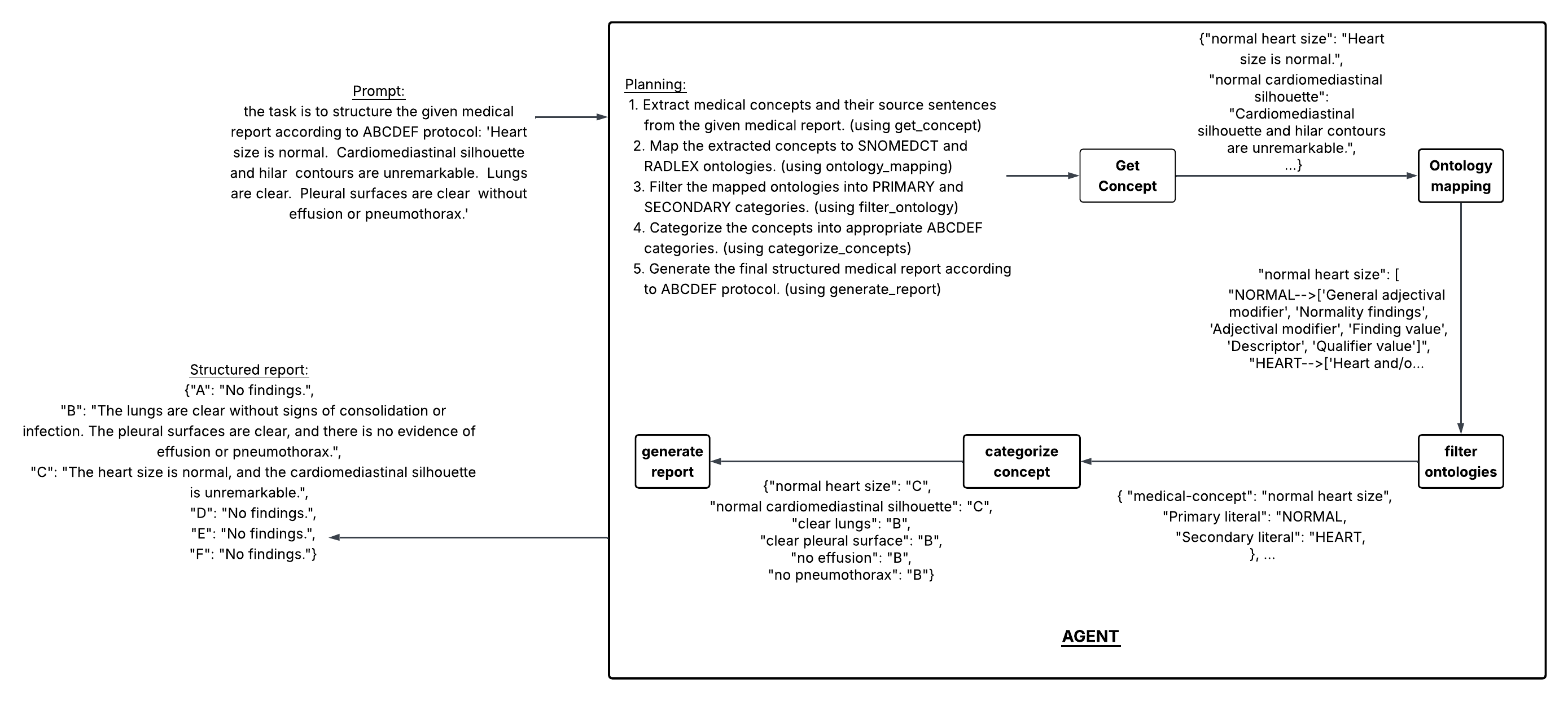}
    \caption{Step by step output of our agent in structuring the free style report according to protocol \cite{Jones_etal_CXR_2025}}
    \label{fig3}
\end{figure}

Figure \ref{fig3} depicts the agent's operational pipeline. Upon receiving an input prompt, the agent first decomposes the task into a sequence of steps and selects the necessary tools. It then executes these steps sequentially, aggregating the outputs from each tool to generate the final structured report.

\section{Results and Analysis}

\subsection{Groundtruth data collection for evaluation}
A key limitation of this study was the absence of large-scale, protocol-aligned structured report data with categorized concepts, which are essential for comprehensive evaluation. To address this gap, we collaborated with radiologists and practicing clinicians to curate a set of ground truth samples for assessing our agent's performance. For this purpose, we extracted relevant findings from the MIMIC-CXR dataset \cite{johnson2019mimic} and developed a dedicated annotation tool to facilitate data collection. The reports were annotated with two key elements: 1) clinically significant concepts extracted from the text, and 2) their respective categorical labels following the defined protocol. In total, 200 ground truth samples were compiled using this method. This strategy facilitated structured validation even in the absence of initially standardized reference data.

\subsection{Concept extraction results}

The performance of the \textit{get concept} tool was evaluated by comparing its extracted concepts against ground-truth annotations from a corpus of radiologist-annotated reports. We framed this task as a multi-label classification problem, where each report may contain multiple relevant or irrelevant concepts from a predefined vocabulary. To quantify performance, we adopted standard multi-label evaluation metrics, including Precision, Recall, F1-score, Subset Accuracy, and Hamming Loss. Due to significant class imbalance in the ground truth data, we also computed macro and weighted averages for Precision, Recall, and F1-score.  Since medical concepts often exhibit lexical variations while conveying the same meaning (e.g., "no effusion" vs. "absence of effusion"), exact string matching was insufficient. Instead, we employed fuzzy matching with similarity thresholds of 80\% and 90\% to align predicted concepts with ground truth annotations. Table \ref{tab1} presents the comparative evaluation of our fine-tuned MedLlama model against leading state-of-the-art LLMs at 80\% and 90\% fuzzy confidence thresholds. Statistical significance of performance differences between models was assessed using McNemar’s test, with corresponding p-values reported for each pairwise comparison.  The results demonstrate that our fine-tuned model not only surpasses comparable-sized models but also outperforms a significantly larger LLM, despite being smaller in scale.

\begin{table}[!t]
    \centering
\caption{Quantitative evaluation for concept extraction}
\label{tab1}
    \begin{tabular}{|>{\centering\arraybackslash}p{0.25\linewidth}|>{\centering\arraybackslash}p{0.11\linewidth}|>{\centering\arraybackslash}p{0.11\linewidth}|>{\centering\arraybackslash}p{0.11\linewidth}|>{\centering\arraybackslash}p{0.11\linewidth}|>{\centering\arraybackslash}p{0.11\linewidth}|>{\centering\arraybackslash}p{0.14\linewidth}|}\hline
 \multicolumn{7}{|c|}{\textbf{Macro and Weighted average metrics for concept extraction @80 fuzzy confidence}}\\\hline
         Model&  Precision&  Recall&  F1 score &  Subset Accuracy& Hamming loss &vs Our Model (p-value)\\\hline
         DeepSeek-R1-Distill-Qwen-7B&  0.78, 0.88&  0.75, 0.82&  0.76, 0.84&  0.13& 0.0049 &0.0007\\\hline
         Qwen2.5-7B&  0.74, 0.86&  0.72, 0.81&  0.73, 0.83&  0.17& 0.0043 &0.0011\\\hline
         Deepseek-R1-70B&  0.88, 0.92&  0.87, 0.90&  0.87, 0.90&  0.15& 0.0037 &0.3613\\ \hline
 \textbf{Our get concept tool}& \textbf{0.87, 0.92}& \textbf{0.86, 0.90}& \textbf{0.86, 0.91}& \textbf{0.28}&\textbf{0.0023} &-\\\hline
 \multicolumn{7}{|c|}{\textbf{Macro and Weighted average metrics for concept extraction @90 fuzzy confidence}}\\\hline
 DeepSeek-R1-Distill-Qwen-7B& 0.72, 0.86& 0.68, 0.71& 0.70, 0.75& 0.07&0.0078 &0.00004\\\hline
 Qwen2.5-7B& 0.69, 0.84& 0.67, 0.75& 0.68, 0.78& 0.17&0.0057 &0.0119\\\hline
 DeepSeek-R1-Distill-Llama-70B& 0.79, 0.88& 0.77, 0.81& 0.77, 0.82& 0.14&0.0069 &0.0519\\\hline
\textbf{ Our get concept tool}& \textbf{0.81, 0.89}& \textbf{0.80, 0.83}& \textbf{0.80, 0.84}& \textbf{0.20}&\textbf{0.0039} &-\\\hline 
    \end{tabular}
 
\end{table}

      

\subsection{Concept categorization results}

A critical component of our agentic framework is the concept categorization tool, which systematically classifies extracted medical concepts into predefined protocol-based categories \cite{Jones_etal_CXR_2025}. We formulate this task as a multiclass classification problem, where the agent must assign each concept to its correct category from a set of categories.

\begin{figure}[!htbp]
    \centering
    \begin{subfigure}[b]{0.32\textwidth}
        \centering
        \includegraphics[width=\linewidth]{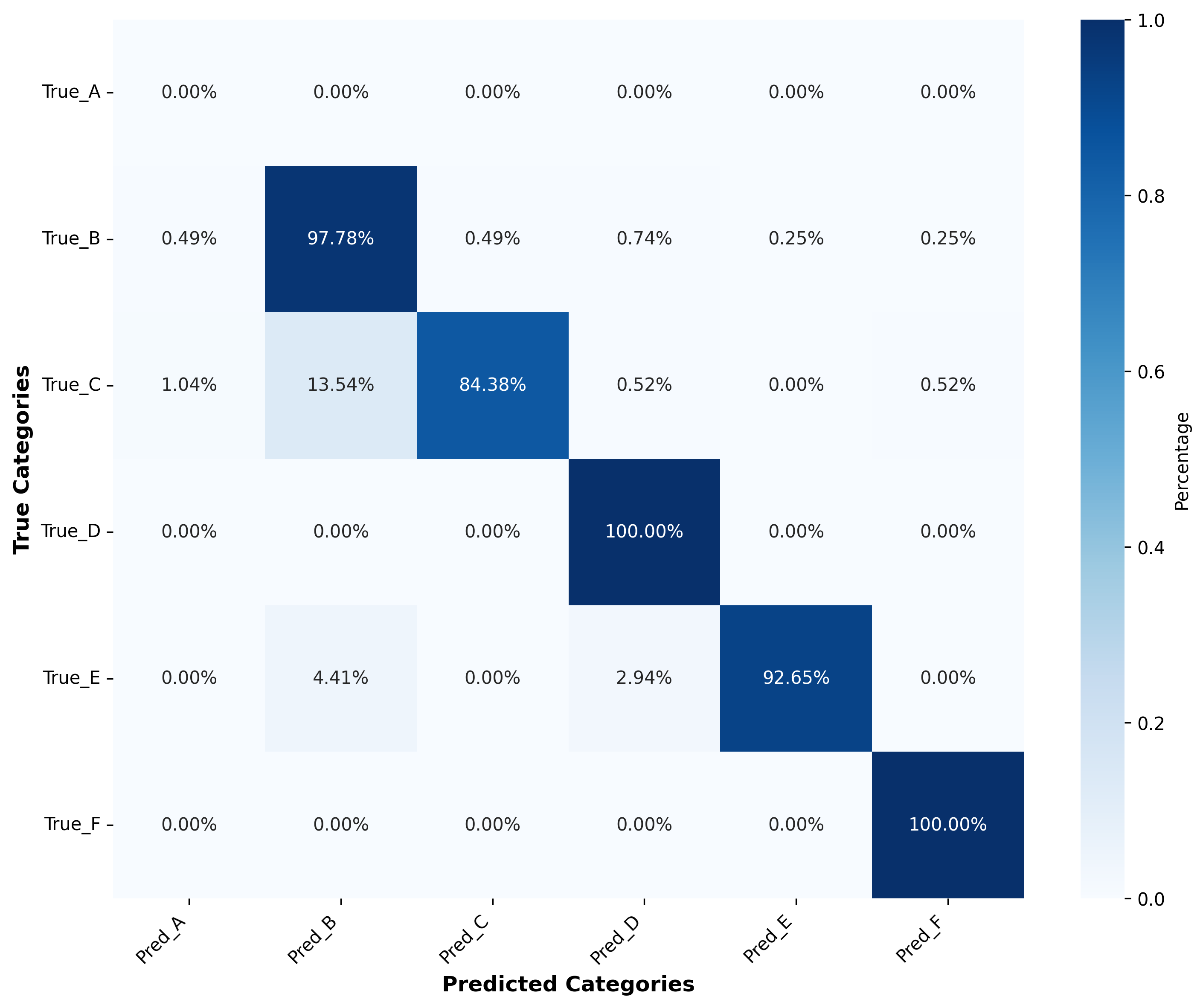}
        \caption{DeepSeek-R1-Distill-Llama-70B}
    \end{subfigure}
   \hfill
    \begin{subfigure}[b]{0.32\textwidth}
        \centering
        \includegraphics[width=\linewidth]{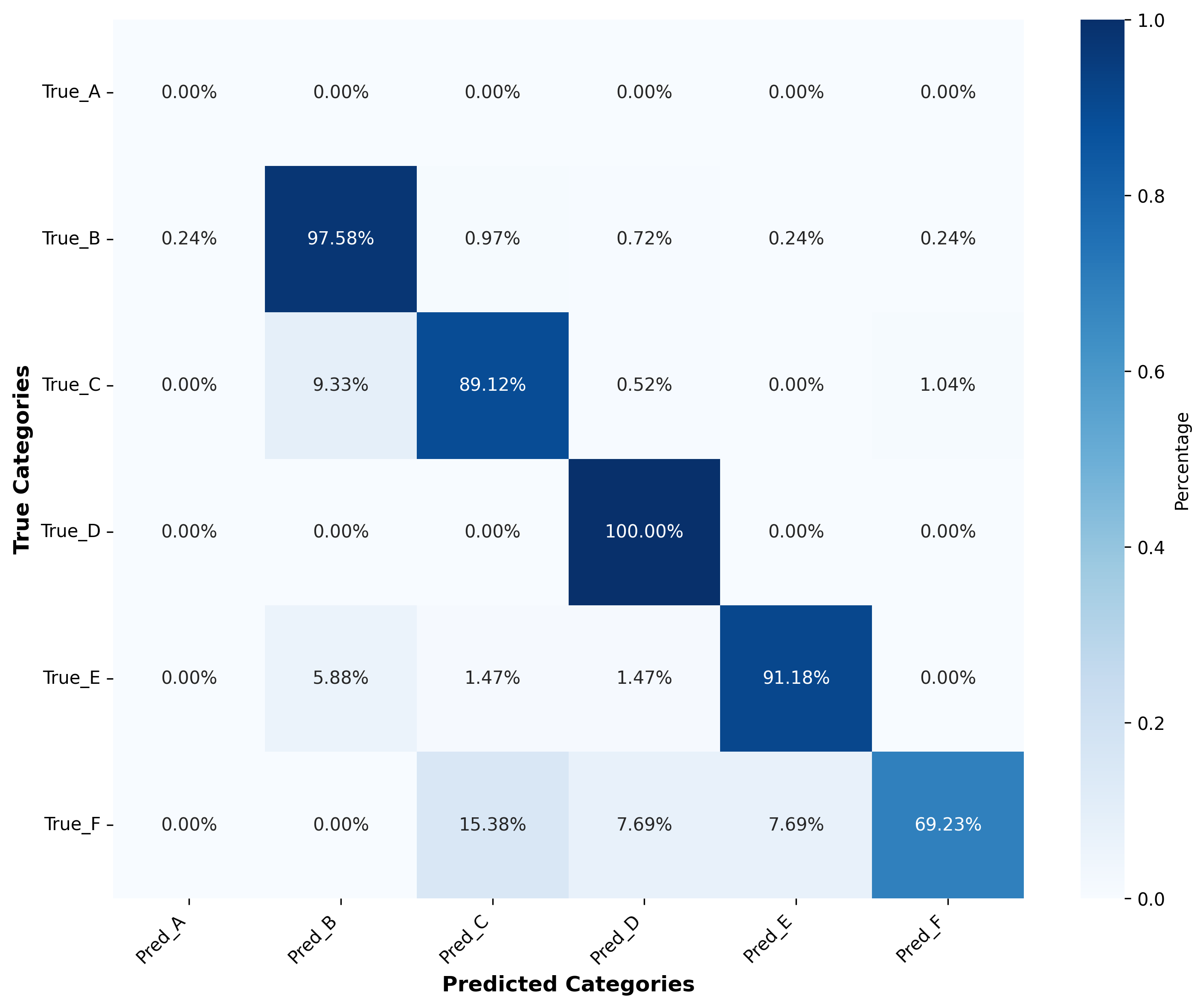}
        \caption{Qwen2.5-72B-Instruct}
    \end{subfigure}
    \hfill
    \begin{subfigure}[b]{0.32\textwidth}
        \centering
        \includegraphics[width=\linewidth]{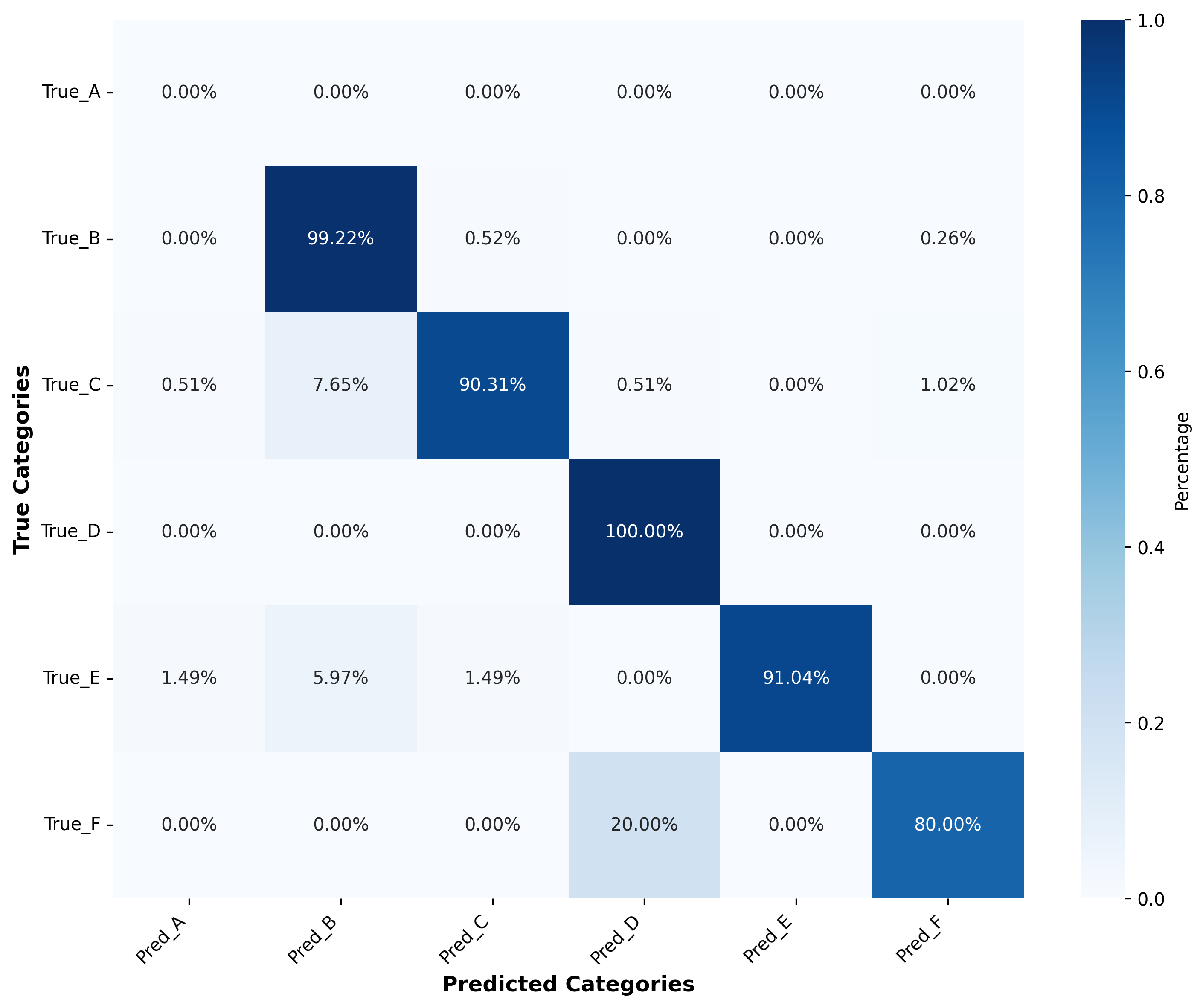}
        \caption{Our Agent}
    \end{subfigure}
    \caption{Confusion matrices over concept categorization task by the respective models}
    \label{fig4}
\end{figure}

\noindent To evaluate performance, we computed standard metrics including Precision, Recall, F1-score, and Jaccard Similarity between predicted and ground truth categorizations. Given the inherent class imbalance, we report both macro and weighted averages (as in Section 4.2). The quantitative results, presented in Table \ref{tab3}, compare our agent against two LLMs, i.e., Deepseek-R1 70B and Qwen 2.5 72B models. Notably, our agent consistently outperforms the base models across all metrics. Figure \ref{fig4} shows the confusion matrix obtained with all the three approaches for the concept categorization task.

\begin{table}
    \centering
\caption{macro and weighted average metrics for concept categorization}
\label{tab3}
    \begin{tabular}{|>{\centering\arraybackslash}p{0.25\linewidth}|>{\centering\arraybackslash}p{0.12\linewidth}|>{\centering\arraybackslash}p{0.12\linewidth}|>{\centering\arraybackslash}p{0.12\linewidth}|>{\centering\arraybackslash}p{0.15\linewidth}|>{\centering\arraybackslash}p{0.15\linewidth}|}\hline
         Model&  Precision&  Recall&  F1 score&  Jaccard Score (IoU) &vs Our Agent (p-values)\\\hline
         DeepSeek-R1-Distill-Llama-70B&  0.93, 0.96&  0.90, 0.93&  0.91, 0.94&  0.88, 0.90 &0.019\\\hline
         Qwen2.5-72B-Instruct&  0.93, 0.96&  0.92, 0.94&  0.92, 0.95&  0.89, 0.92 &0.261\\\hline
         \textbf{Our agent}&  \textbf{0.95, 0.97}&  \textbf{0.94, 0.96}&  \textbf{0.94, 0.96}&  \textbf{0.92, 0.94} &-\\ \hline
    \end{tabular}

\end{table}

\subsection{Evaluating the generated reports}
To assess the quality of the generated radiology reports, we adopted an evaluation framework inspired by the work of Dom Marshall et al. \cite{sharma2024cxr}. We introduced two key scoring metrics: (1) an Accuracy Score, which measures the clinical correctness of the generated findings against the original MIMIC-CXR reference reports, and (2) a Structural Score, which evaluates the degree to which the generated reports adhere to the predefined reporting protocol. These metrics provide a comprehensive assessment of both the medical validity and standardized formatting of the outputs, which are represented in Table \ref{tab4}.

\begin{table}
\centering
\renewcommand{\arraystretch}{1.1} 
\caption{Scoring criteria for report quality}
\label{tab4}
\begin{tabular}{>{\centering\arraybackslash}p{0.1\linewidth}p{0.45\linewidth}>{\raggedright\arraybackslash}p{0.45\linewidth}}
\toprule
\textbf{Score} & \textbf{Definition for accuracy scoring} &\textbf{Definition for structure scoring}\\
\midrule
5 &  Perfect Match: All key details are present, and there is no missing, extraneous, or fabricated (hallucinated) information. &Perfect Structure: All specified Key Concepts are present as distinct sections/headers.\\
4 & Generally Accurate with Minor Omissions: It may be missing a few minor details that do not impact the overall meaning or interpretation. No hallucinations. &Good Structure: All Key Concepts are present and correctly populated, but there might be minor deviations. \\
3 & Key Details Present but with Minor Hallucinations: The Generated Report captures the main points but has significant omissions of secondary details. &Acceptable Structure: Most Key Concepts are present, but one or two may be missing or merged.\\
2 & Missing Key Details or Contradictory Information: The Generated Report fails to mention one or more key details from the groundtruth.  &Poor Structure: Multiple Key Concepts are missing. Information is disorganized, placed under incorrect headings, or lumped together.\\
1 & Critically Flawed or Dangerous: The Generated Report has major omissions of critical information or contains significant, dangerous hallucinations. &No Structure: The report is a free-form block of text. It completely ignores the given  protocol-driven structure.\\
\bottomrule
\end{tabular}
\end{table}

\noindent To validate the clinical quality of our agent's outputs, we conducted a evaluation study involving certified radiologists and clinician. The evaluation procedure required radiologists and clinician to assess 50 samples of our agent-generated reports by comparing them against corresponding reference reports from the MIMIC-CXR dataset (distinct from the data used in Sections 4.2 and 4.3). The assessment was conducted using the scoring criteria specified in Table \ref{tab4}.  The mean of accuracy score and structure scores are as shown in Table \ref{tab6}.  

\begin{table}[!h]
    \centering
\caption{Radiologist and Clinician evaluation of agent generated reports}
\label{tab6}
    \begin{tabular}{|c|c|c|}\hline
         Method&  Average Accuracy score&Average Structure score\\\hline
         Clinician&  4.64&4.59\\\hline
 Panel of two Radiologists& 4.52&4.48\\\hline
    \end{tabular}
\end{table}
    
The complete agent framework was deployed on Intel's Gaudi2 AI accelerator, leveraging its specialized architecture for efficient deep learning inference.  We conducted rigorous timing evaluations to quantify the agent's operational efficiency across its core functionalities. As documented in Table \ref{tab7}, these measurements encompass both the integrated agent pipeline and its individual component modules. The agent supports four primary transformation tasks: (1) conversion of unstructured clinical narratives to standardized reports, (2) generation of structured reports from extracted medical concepts, (3) organization of raw medical concepts into structured representations, and (4) direct extraction of structured concepts from unstructured text.


\begin{table}[H]
    \centering
\caption{Inference time of each tool and overall agentacross multiple tasks(in seconds)}
\label{tab7}
    \begin{tabular}{|>{\centering\arraybackslash}p{0.18\linewidth}|>{\centering\arraybackslash}p{0.11\linewidth}|>{\centering\arraybackslash}p{0.11\linewidth}|>{\centering\arraybackslash}p{0.11\linewidth}|>{\centering\arraybackslash}p{0.11\linewidth}|>{\centering\arraybackslash}p{0.11\linewidth}|>{\centering\arraybackslash}p{0.11\linewidth}|>{\centering\arraybackslash}p{0.1\linewidth}|}\hline
         Tasks&   Planning time&get concept&  ontology mapping&  ontology filtering&  concept categorization&  report generation& Overall\\\hline
         freestyle report-to-structured report&   46&1&  56&  132&  25&  20& 280\\\hline
         freestyle report-to-structured concepts&   45&1&  55&  130&  22&  -& 254\\\hline
         raw concepts-to-structured report&   45&-&  53&  134&  23&  22& 277\\\hline
         raw concept-to-structured concept&   45&-&  53&  140&  23&  -& 261\\\hline
 freestyle report-to-structured report with check cache enabled& 45& 1& -& -& -& 20&66\\\hline
    \end{tabular}

\end{table}

\section{Conclusion}


In conclusion, our work introduces a novel protocol-driven, ontology-aware AI framework that outperforms existing models in radiology report structuring, achieving superior clinical correctness while maintaining diagnostic accuracy. This agent serves as an efficient tool for radiologists, enabling them to generate fully structured reports by simply providing key impressions and findings - significantly reducing documentation effort while ensuring standardization and without replacing expert judgment.  The framework’s modality-agnostic design allows for seamless expansion beyond chest radiography to other imaging modalities, enhancing its versatility across radiology workflows. Looking ahead, integration with medical images and automated suggestion systems could further augment its utility, while optimizations in model compression and hardware acceleration may enable real-time clinical deployment. These advancements position the framework for broad adoption across healthcare institutions with varying reporting standards and resources, ultimately improving documentation consistency and workflow efficiency.

\begin{credits}
\subsubsection{\ackname} We acknowledge Intel CSR grant to IITM Pravartak which enabled the compute requirement.

\subsubsection{\discintname}
The authors have no competing interests to declare that are
relevant to the content of this article.
\end{credits}
%
%
%
\bibliographystyle{splncs04}
\bibliography{thebibliography}
%




\end{document}